\documentclass[xcolor=table,10pt,twocolumn,letterpaper]{article}

\usepackage[pagenumbers]{cvpr} 
\usepackage{graphicx}
\usepackage{amsmath}
\usepackage{amssymb}
\usepackage{booktabs}
\usepackage{float}
\usepackage{multirow}
\usepackage[normalem]{ulem}
\usepackage[table,xcdraw]{xcolor}
\usepackage[linesnumbered,ruled,vlined]{algorithm2e}
\usepackage{makecell}

\usepackage[pagebackref,breaklinks,colorlinks]{hyperref}

\usepackage[capitalize]{cleveref}
\crefname{section}{Sec.}{Secs.}
\Crefname{section}{Section}{Sections}
\Crefname{table}{Table}{Tables}
\crefname{table}{Tab.}{Tabs.}

\def\cvprPaperID{23} 
\def\confName{CVPR}
\def\confYear{2023}

\begin{document}

\title{PointCLIMB: An Exemplar-Free Point Cloud Class Incremental Benchmark}


\author{${\text{Shivanand Kundargi}}^{*}$, ${\text{Tejas Anvekar}}^{*}$, Ramesh Ashok Tabib, Uma Mudenagudi\\
Center of Excellence in Visual Intelligence (CEVI), KLE Technological University.\\
Vidya Nagar, Hubballi, Karnakata, India.\\
\tt\small shivanandkundargi992@gmail.com, anvekartejas@gmail.com, ramesh\_t@kletech.ac.in, uma@kletech.ac.in}

\maketitle


\begin{abstract}
     Point clouds offer comprehensive and precise data regarding the contour and configuration of objects. Employing such geometric and topological 3D information of objects in class incremental learning can aid endless application in 3D-computer vision. Well known 3D-point cloud class incremental learning methods for addressing catastrophic forgetting  generally entail the usage of previously encountered data, which can present difficulties in situations where there are restrictions on memory or when there are concerns about the legality of the data. Towards this we pioneer to leverage exemplar free class incremental learning on Point Clouds. In this paper we propose PointCLIMB: An exemplar Free Class Incremental Learning Benchmark. We focus on a pragmatic perspective to consider novel classes for class incremental learning on 3D point clouds. We setup a benchmark for 3D Exemplar free class incremental learning. We investigate performance of various backbones on 3D-Exemplar Free Class Incremental Learning framework. We demonstrate our results on ModelNet40 dataset. 
\end{abstract} 




. 
 
\section{Introduction}
Point cloud analysis with pioneering works exploring global \cite{pointnet} and local \cite{dgcnn,pointnetpp,PointMLP,PRA-NET} geometries has become an increasingly popular approach for understanding 3D objects and environments, with many potential applications\cite{wang2019applications} in real-time settings. Considering the real-time applications of point cloud analysis, there is always a new incoming stream of data available to the learner. Comparing this realistic scenario to human cognition, humans leverage their existing knowledge and build upon it when learning new things, rather than starting from scratch every time. In real-world, for the tasks such as autonomous driving and robotic applications, the data is not static but rather accumulates over time. Therefore, mimicking the human cognition there is a dire need for models built upon such real world applications to learn incrementally as new data is added, rather than retraining the entire model from scratch each time. This approach is known as class incremental learning(a subset of continual or life long learning)\cite{masana2022class}, which allows the model to adapt to changes in the data distribution over time, while retaining knowledge learned from previous data\cite{pfulb2018catastrophic}.

Class-incremental learning has been investigated to a certain extent in 2D(image data) realm\cite{rebuffi2017icarl,li2017learning,smith2021always}, its exploration on 3D data(point cloud data) has been relatively limited. There have been initial works towards mitigating catastrophic forgetting in point cloud class incremental learning 3D-FSCIL\cite{chowdhury2022few}, I3DOL\cite{dong2021i3dol}. these methods have an unfortunate lacunae: They require extensive memory bank for replay of point cloud data from previous task. The problem of memory constraints in class-incremental learning is a matter of great concern in computer vision for two primary reasons. Firstly, numerous applications with point cloud analysis operate on devices with limited memory capacity, rendering the issue of memory consumption becomes a crucial consideration \cite{liu2021lifelong,soma2020fast}. Secondly, many 3D-computer vision applications acquire data that is subject to legal restrictions\cite{zhu2020private,das2017assisting,beaulieu2019privacy}, making storage a difficult and often infeasible task. These caveats lead us to ask How can 3D-computer vision systems incrementally incorporate new geometric information without storing previous data. 

Towards modeling 3D-EFCIL(3D Exemplar Free Class incremental learning), we propose PointCLIMB: An Exemplar-Free Point Cloud Class Incremental Benchmark. We summarize our contributions as follows
\begin{itemize}
    \item We are the first to model 3D-EFCIL and benchmark the results on Modelnet40 dataset\cite{wu20153d}.
    \item We are the first to investigate the importance of backbone(feature extractor) in 3D-EFCIL.
    \item We propose to employ a weighted knowledge distillation loss towards mitigating catastrophic forgetting.
    \item We propose to model a pragmatic approach to benchmark 3D-EFCIL, contrast to other works in point cloud class-incremental learning which never focus on pragmatic approach considering novel data arrival.
\end{itemize}

\begin{figure*} [ht]
    \centering
    \captionsetup{type=figure}
    \includegraphics[width=\linewidth]{ 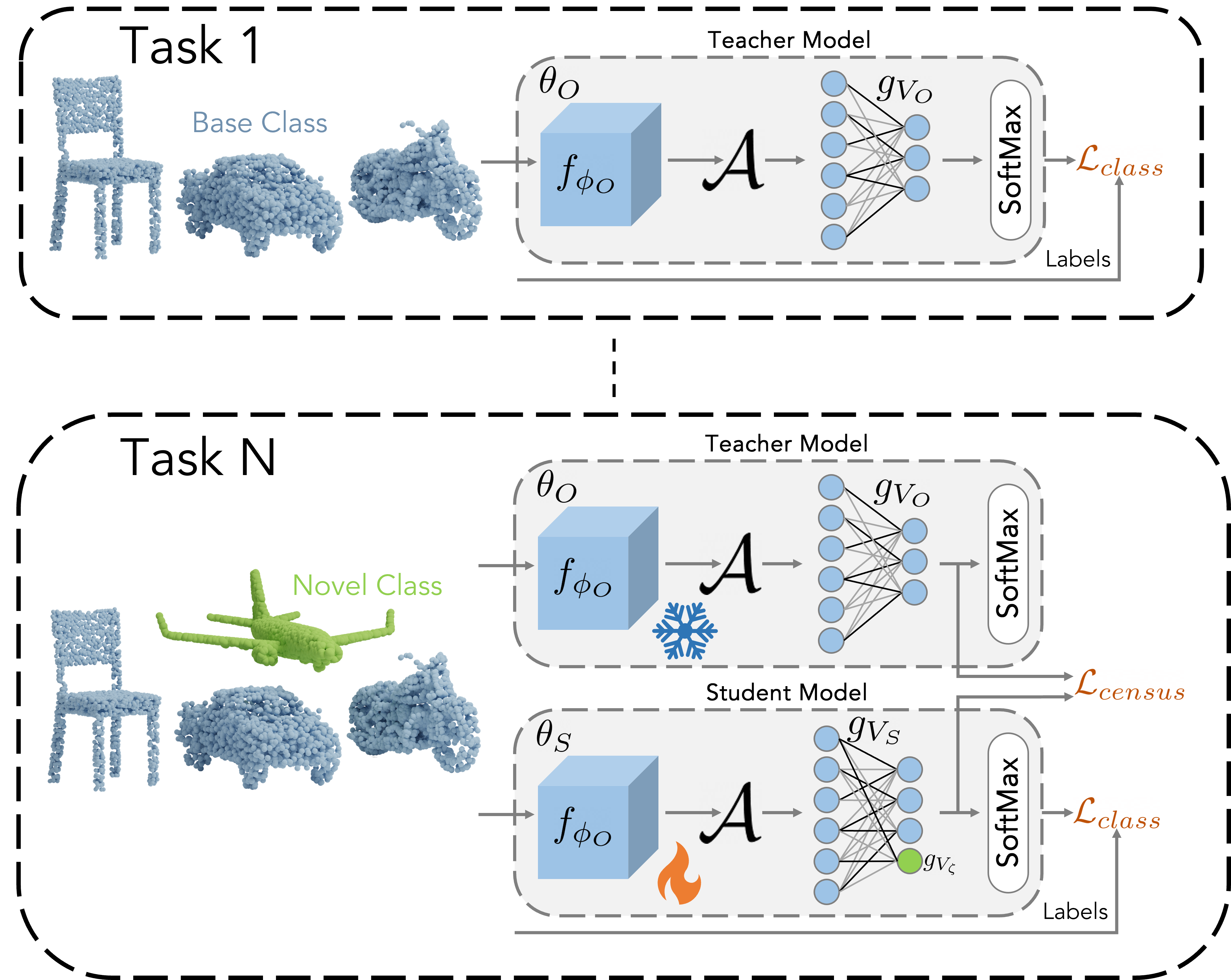}
    \captionof{figure}{The process of network optimization in \textbf{PointCLIMB} can be illustrated through two tasks. The first, base task involves training a teacher model using a feature extractor $f$ with parameters $\phi_{O}$ and a linear classifier $g$ with parameters $V_{O}$. The operation $\mathcal{A}$, which can be mean, max or sum, is applied to make the output symmetric. In the second, class incremental novel task N; a student model is introduced when a novel task N arrives. The weights of the student model are initialized by copying the weights of the teacher model. Specifically, $g(;V_{S}) = g(;V_{O}) \cup g(;V_{\zeta})$, where $g(;V_{\zeta})$ represents the weights associated with the novel class. The teacher model is kept frozen during this process. To mitigate the issue of catastrophic forgetting, we use census knowledge distillation loss, which compares the logits of the teacher and student models.}
    \label{fig:teaser}
\end{figure*}

\section{Related Works}
\noindent \textbf{Point Cloud Analysis}
With the recent emergence of Lidar sensors, numerous studies have been conducted on directly classifying 3D point cloud objects. voxel and multiview based methods \cite{mvcnn, xu2021voxel} were the works carried out initially towards point cloud analysis. PointNet\cite{pointnet}, was the first to employ multi-layer perceptron (MLP) networks to interpret 3D point clouds. Addressing the limitation of considering only global topological information, PointNet++\cite{pointnetpp} was proposed which considers the deep hierarchical features for pointcloud processing. DGCNN\cite{dgcnn} proposed Edge-Conv a graph based dynamic convolution for exploiting local geometry of a point cloud. PointMLP\cite{PointMLP}, a purely residual MLP network that achieves high performance without the need for complex local geometric extractors. Despite its simplicity, PointMLP performs remarkably well and is highly competitive with more sophisticated models. These local and global topology aware methodologies never explore learning similarities between region of point cloud.Towards addressing the aforementioned challenge Point Relation-Aware Network (PRA-Net)\cite{PRA-NET} was proposed, which is composed of an Intra-region Structure Learning (ISL) module and an Inter-region Relation Learning (IRL) module. The ISL module has the ability to dynamically integrate local structural information into point features, while the IRL module can effectively capture inter-region interactions using a differentiable region division method and a representative point-based technique, which is both adaptable and efficient. Hence methods that explore both inter-region Relation and Intra-region Structure Learning which extract superior geometric features might strengthen the backbone for Class Incremental Learning on point clouds

\noindent \textbf{Incremental Learning}
Task Incremental, Domain Incremental and Class Incremental are the three main categories in continual learning. The issue of catastrophic forgetting has been widely recognized for many years, with evidence dating back to the 1980s when \cite{mccloskey1989catastrophic} demonstrated that algorithms trained with backpropagation were prone to this phenomenon. Subsequent research by \cite{ratcliff1990connectionist} corroborated these findings and extended them to a broader range of tasks trained using backpropagation. A comprehensive review of early attempts to address catastrophic forgetting is provided by \cite{french1999catastrophic}.There have been many subsequent works in 2D continual learning whereas in 3D realm, the continual learning framework has been underexplored. 3D-FSCIL\cite{chowdhury2022few} explored the few shot aspect of class incremental learning on point clouds. They propose microshapes to mitigate catastrophic forgetting. The existing 3D class-incremental learning methods store exemplars or microshapes  data securities and memory.I3DOL\cite{dong2021i3dol} proposes to mitigate the issue of catastrophic forgetting that can result from the presence of redundant geometric information,towards this an attention mechanism that is sensitive to geometric properties has been introduced. This mechanism quantifies the significance of local geometric structures and identifies distinctive 3D geometric attributes that make substantial contributions to incremental learning of classes.Though aforementioned 3D Class-Incremental learning methods are well established they make extensive replay of previous data for training incrementally.

\noindent \textbf{Exemplar-Free Class-Incremental learning} According to recent reviews of class incremental learning (CIL), the majority of methods aimed at mitigating catastrophic forgetting incorporate techniques that involve replaying samples from past classes\cite{End-to-End-Incremental-Learning}.which have data privacy restrictions or storage limitations\cite{venkatesan2017strategy}. Towards addressing the aforementioned challange  Learning Without Forgetting\cite{li2017learning} was proposed as knowledge distillation method for class-incremental learning set-up. \cite{kirkpatrick2017overcoming} weight and data regularisation based methods were proposed to mitigate catastrophic forgetting without storing exemplars.

\section{PointCLIMB}

In this section, we present a practical scenario-based assessment of the 3D-EFCIL model, dubbed as PointCLIMB. Additionally, we propose robust benchmark models, conduct comprehensive experiments, and elucidate the rationale behind the selection of backbone architectures.

Exemplar free class incremental learning (3D-EFCIL) is quintessential yet an under-explored paradigm in the realm of point cloud analysis. To facilitate benchmarking in realistic scenarios for 3D-EFCIL, this paper introduces \textbf{PointCLIMB}: An Exemplar Free \textbf{Point} Cloud \textbf{CL}ass \textbf{I}ncre\textbf{M}ental \textbf{B}enchmark. PointCLIMB investigates the need for point cloud backbones that extracts superior geometric features capable of mitigating the challenge of catastrophic forgetting in realistic 3D-EFCIL settings. We propose to employ Census; a weighted knowledge distillation loss between old and new class logits of point cloud backbones. Census is dynamic to incremental classes that enhances the performance of point cloud backbones in 3D-EFCIL. We employ veristic task sampler which mimicks the natural way of sampling the tasks to be learnt incrementally in 3D-EFCIL setting.

        Incremental learning problem $\mathcal{T}$ consists of sequence of $m$ tasks:
        \begin{equation}
            \label{eq:3defcil-setting}
            \mathcal{T} = [(C^{1},D^{1}), (C^{2},D^{2})\text{, ... ,}(C^{m},D^{m})]
        \end{equation}

    \subsection{PointCLIMB Settings}

        \newcommand\mycommfont[1]{\footnotesize\ttfamily\textcolor{blue}{#1}}
        \SetCommentSty{mycommfont} 
        
        \begin{algorithm}[h]
            
            \DontPrintSemicolon
            \SetKwData{tc}{tc}
            \SetKwData{h}{high}
            \SetKwData{l}{low}
            \SetKwData{base}{base}
            \SetKwData{append}{append}
            \SetKwData{TL}{TL}
            
            \SetKwData{condition}{condition}
            \SetKwData{classes}{classes}
            
            \SetKwInOut{Input}{Input}
            \SetKwInOut{Output}{Output}
            
            \Input{\tc, \l, and \h.} \tcc{\tc is total number of classes, \l represents least no of categories present in a task, and \h represents maximum no of categories present in a task.}
            \Output{The lists $\mathcal{T}$.} \tcp{$\mathcal{T}$ is task list for continual learning.}
            
            \BlankLine
            \classes $\gets$ arrange(0,\tc-1);
            shuffle(\classes)\;
            
            \TL $\gets$ [];
            $\mathcal{T} \gets$ []\;
            \condition $\gets 0$
            
            \While{\tc $\neq$ 0 \textbf{and} \condition $\geq$ 0}{
            \BlankLine
            \base $\gets$ RANDOMINT(\l, \h)\;
            \condition $:=$ \tc - \base\;
            \BlankLine
            \If{\condition $\leq$ 0}{
            \TL.\append(\tc)\;
            \textbf{break}\;
            }
            \Else{
            \TL.\append(\base)\;
            }
            \tc $:=$ \condition\;
            }
            
            s $\gets$ 0\;
            \For{i $\gets$ 0 \KwTo length(\TL)}{
            $\mathcal{T}$.\append $\Big($\classes$\big[$\TL[s]:\TL[s]+\TL[i]$\big]$$\Big)$\;
            s $:=$ \TL[i]\;
            }
            
            \BlankLine
            \Return{ $\mathcal{T}$};
            
            \caption{Veristic Task Sampler: A pragmatic sampler mimicking the paradigm of class incremental learning}
            \label{alg:veristic}
        \end{algorithm}
        The veristic task sampler is used to model $\mathcal{T}$ by selecting tasks based on a naturalistic setting that imitates the paradigm of novel data arrival in Class Incremental learning best explained in Algorithm  \ref{alg:veristic}. Where each task $t$ is represented by a set of classes $C^{t} = \{c^{t}_{1},c^{t}_{2}\text{, ... ,}c^{t}_{m^{t}}\}$ and the training data $D^{t}$. We use $M^{t}$ to represent the total number of classes in all tasks up to and including task $t: M^t = \sum_{i=1}^{t} C^i$. The 3D class-incremental problem in which $D^t = \{(p_{1},y_{1}), (p_{2},y_{2})\text{, ... ,}(p_{l^{t}},y_{l^{t}})\}$, where $p$ is point cloud with $n$ points such that $p \in \mathbb{R}^{n \times 3}$. During training for task $t$, the learner only has access to $C^t, D^t$ where as during inference the evaluation is done for the union of all previous tasks $\bigcup_{i=1}^{t} C^i,D^i$. For instance if we encounter task $t=2$, the learner has access to $(C^2,D^2)$ where as evaluation is done for $\{(C^1,D^1), (C^2,D^2)\}$. Towards modeling 3D-EFCIL problem setting; 1: we don not allow class overlaps between tasks (i.e, $C^i \cap C^j = $\O  if $i \neq j$), 2: we do not maintain any coreset(exemplars) of previous task $t-1$ for training the current task $t$.

        We consider incremental learners, the teacher model $O(p, \theta_{O})$ parameterized by weights $\theta_{O}$ and the student model $S(p, \theta_{S})$ parameterized by weights $\theta_{S}$ to indicate the output logits of the network on input $p$. We further split the neural network in a feature extractor $f$ with weights $\phi$ and linear classifier $g$ with weights $V$ according to $O(p, \theta_{O}) = g\big(f(p;\phi_{O}) ;V_{O} \big)$ and $S(p, \theta_{S}) = g\big(f(p;\phi_{O}) ;V_{S} \big)$. The student model's weights are designed such that $V_{S} = V_{O} \cup V_{\zeta}$ where $V_{\zeta}$ is novel task specific parameters as shown in Figure \ref{fig:teaser}. We use $({\hat{y}_{O};\tau})= \sigma(O(p;\theta_{O}), \tau)$ and $({\hat{y}_{S};\tau})= \sigma(S(p;\theta_{S}), \tau)$ to identify teacher and student network predictions, where $\sigma(,\tau)$ indicates the softmax functions with temperature $\tau$.\\

    \subsection{Influence of Backbones on PointCLIMB}
        To investigate the importance of superior and robust feature extraction in PointCLIMB, we train point cloud class-incremental classifiers as per PointCLIMB Settings. For our case study we select three types of open-source state-of-the-art backbones.
        \begin{itemize}
            \item Global per-point based PointNet\cite{pointnet}.
            \item Local neighbourhood (Intra-region) based PoinetNet++\cite{pointnetpp}, DGCNN\cite{dgcnn}, PointMLP\cite{PointMLP}.
            \item Intra-region structure aware and Inter-region relation aware PRA-Net\cite{PRA-NET}.
        \end{itemize}
        Backbones act as a feature extractor whose features are used for knowledge distillation in exemplar free class-incremental learning Methods. None of the previous works on 3D point cloud class incremental learning focuses on the importance of feature extraction. We investigate the significance of feature extractor by evaluating different state of the art classification networks as feature extractor on PointCLIMB settings. Our findings infer that, backbones that incorporate both  Intra-region structure and Inter-region relation aware properties perform significantly better in PointCLIMB settings compared to other types as depicted in Table \ref{tab:all-loss}.
        The reason for this may be the extracted features of PRA-Net\cite{PRA-NET} best describe the topology, similarity, proximity, and symmetry of a point cloud when compared with graph based semantic features of DGCNN. Another reason is due to the incorporation of gated units in IRL blocks of PRA-Net which resembles highway-connection classifier networks (HCNs)\cite{highway}. We conclude that networks with gated unit incorporates high stability as explained by \cite{highway} in 2D realm.
        
    \subsection{Knowledge Distillation}
        The process of Knowledge Distillation in 3D-EFCIL involves the transfer of geometric and topological knowledge from a previously trained model (known as the teacher model: $O(p,\theta_{O})$) to a new model (referred to as the student model: $S(p,\theta_{S})$). This transfer of knowledge aims to enable the student model to accurately classify new point cloud categories, while still retaining its ability to classify the categories it was previously trained on. 
        
        \noindent \textbf{Learning Without Forgetting} (LwF) enables the student model to learn new tasks without forgetting the knowledge it has already acquired, by leveraging the distillation of knowledge from the teacher model and can be expressed as:
        \begin{equation}
        \label{eq:lwf}
            \mathcal{L}_{LwF} = \lambda \mathcal{L}_{distill} + \mathcal{L}_{class}
        \end{equation}
        where $\mathcal{L}_{distill}$ is the distillation loss term, which measures the difference between the output probabilities of the teacher model and the student model, and is given by:
        \begin{equation}
        \label{eq:dist}
            \mathcal{L}_{distill} = - \frac{1}{N} \sum_{i=1}^{n} (\hat{y}^{i}_{O};\tau) \log{\big((\hat{y}^{i}_{S};\tau)\big)}
        \end{equation}
        Here, $\hat{y}^{i}_{O}$ represents the output probability of the teacher model for the i-th input sample, and $\hat{y}^{i}_{S}$ represents the output probability of the student model for the same input sample with temperature $\tau$.
        The second term, $\mathcal{L}_{class}$, is the classification loss term, which measures the deviation of the student model's output probabilities from the true labels of the current task, and is given by:
        \begin{equation}
        \label{eq:class}
            \mathcal{L}_{class} = - \frac{1}{N} \sum_{i=1}^{n} y^i \log{\hat{y}^{i}_{S}}
        \end{equation}
        Here, $y^i$ represents the true label of the i-th input sample.
        Finally, $\lambda$ is a hyperparameter that balances the relative importance of the distillation and classification loss terms.\\

        \noindent \textbf{Census Knowledge Distillation} is an improved variant of LwF that dynamically adjusts the weight of the distillation loss term for every new increment of task $t$, based on the current count of classes in task $\eta$ and the number of tasks elapsed $T$. This dynamic weight helps to ensure robustness towards catastrophic forgetting and can be expressed as
         \begin{equation}
            \label{eq:lcensus}
            \mathcal{L}_{census} = (\eta * T) \mathcal{L}_{distill}
        \end{equation} 

        where $\mathcal{L}_{distill}$ is briefed in Eq. \ref{eq:dist} Allocating dynamic weights to the knowledge distillation loss based on the arrival of new tasks enhances the importance of geometric features associated with the newly arrived classes. This approach can help mitigate the issue of task recency bias, ultimately leading to improved performance. To investigate the significance of employed Census knowledge distilation loss, we train PointNet~\cite{pointnet}, PointNet++~\cite{pointnetpp}, DGCNN~\cite{dgcnn}, PointMLP~\cite{PointMLP}, and PRA-Net~\cite{PRA-NET} on PointCLIMB settings and compare with LwF~\cite{li2017learning} as depicted in Table~\ref{tab:all-loss}.

\section{Experiments}
\noindent \textbf{Training pipeline} of \textbf{PointCLIMB} is based on a fine-tuning strategy, which involves two steps. Firstly, a base task is trained on a backbone architecture using the cross-entropy loss given by Eq. \ref{eq:class}. We refer to this model as the teacher model.Subsequently, when an incremental task containing a novel class is presented, we compute a weighted knowledge distillation loss between the logits of the weight-frozen teacher model and the weight-shared student model, as shown in Eq. \ref{eq:lcensus}. This loss is combined with the cross-entropy loss to train the novel set of classes. In other words, the student model learns from the teacher model while retaining the knowledge of the previous tasks, by jointly minimizing the cross-entropy loss and the weighted knowledge distillation loss. This approach ensures that the student model adapts to the new task while preventing catastrophic forgetting of the previously learned tasks.\\

\useunder{\uline}{\ul}{}
\begin{table}[t]
\centering
\caption{Performance of different backbone's with PointCLIMB on ModelNet40 dataset. \textit{\textbf{Joint}} depicts Upper bound and \textit{\textbf{FT}} represents Fine-Tuning approach. We demonstrate supremacy of census over other knowledge distillation loss for each backbone on 3D-EFCIL. We represent our findings by 1) best by \textbf{\uline{bold underline}} and 2) second best by \textbf{bold} accuracy values.}
\label{tab:all-loss}
\resizebox{\linewidth}{!}{%
\begin{tabular}{rcccccc}
\hline \hline
 
\textbf{Methods} &
  \textbf{Loss} &
  \textbf{20} &
  \textbf{5} &
  \textbf{5} &
  \textbf{5} &
  \textbf{5} \\ \hline 
 &
  \cellcolor[HTML]{DAE8FC}\textbf{\textit{Joint}} &
  \cellcolor[HTML]{DAE8FC}94.77 &
  \cellcolor[HTML]{DAE8FC}92.11 &
  \cellcolor[HTML]{DAE8FC}86.98 &
  \cellcolor[HTML]{DAE8FC}85.43 &
  \cellcolor[HTML]{DAE8FC}84.67 \\
 &
  \textbf{\textit{FT}} &
  94.10 &
  02.65 &
  01.61 &
  00.31 &
  00.24 \\
 &
  \textbf{LwF} &
  {\ul \textbf{95.25}} &
  \textbf{53.37} &
  \textbf{04.92} &
  \textbf{02.62} &
  \textbf{00.32} \\
\multirow{-4}{*}{\textbf{PointNet}~\cite{pointnet}} &
  \textbf{Census} &
  \textbf{94.01} &
  {\ul \textbf{67.77}} &
  {\ul \textbf{49.25}} &
  {\ul \textbf{26.95}} &
  {\ul \textbf{18.47}} \\
 &
  \cellcolor[HTML]{DAE8FC}\textbf{\textit{Joint}} &
  \cellcolor[HTML]{DAE8FC}93.17 &
  \cellcolor[HTML]{DAE8FC}92.76 &
  \cellcolor[HTML]{DAE8FC}87.34 &
  \cellcolor[HTML]{DAE8FC}85.29 &
  \cellcolor[HTML]{DAE8FC}85.32 \\
 &
  \textbf{\textit{FT}} &
  92.33 &
  01.23 &
  01.28 &
  00.58 &
  00.24 \\
 &
  \textbf{LwF} &
  \textbf{92.51} &
  \textbf{61.34} &
  \textbf{12.63} &
  \textbf{07.32} &
  \textbf{01.86} \\
\multirow{-4}{*}{\textbf{PointNet++}~\cite{pointnetpp}} &
  \textbf{Census} &
  {\ul \textbf{93.39}} &
  {\ul \textbf{71.24}} &
  {\ul \textbf{56.26}} &
  {\ul \textbf{44.09}} &
  {\ul \textbf{29.09}} \\
 &
  \cellcolor[HTML]{DAE8FC}\textbf{\textit{Joint}} &
  \cellcolor[HTML]{DAE8FC}95.44 &
  \cellcolor[HTML]{DAE8FC}94.62 &
  \cellcolor[HTML]{DAE8FC}88.13 &
  \cellcolor[HTML]{DAE8FC}86.15 &
  \cellcolor[HTML]{DAE8FC}85.87 \\
 &
  \textbf{\textit{FT}} &
  94.67 &
  01.88 &
  01.23 &
  00.63 &
  00.68 \\
 &
  \textbf{LwF} &
  {\ul \textbf{95.50}} &
  \textbf{61.21} &
  \textbf{06.74} &
  \textbf{01.99} &
  \textbf{00.93} \\
\multirow{-4}{*}{\textbf{PointMLP}~\cite{PointMLP}} &
  \textbf{Census} &
  \textbf{95.25} &
  {\ul \textbf{64.39}} &
  {\ul \textbf{52.30}} &
  {\ul \textbf{27.49}} &
  {\ul \textbf{28.68}} \\
 &
  \cellcolor[HTML]{DAE8FC}\textbf{\textit{Joint}} &
  \cellcolor[HTML]{DAE8FC}96.52 &
  \cellcolor[HTML]{DAE8FC}93.93 &
  \cellcolor[HTML]{DAE8FC}88.49 &
  \cellcolor[HTML]{DAE8FC}87.96 &
  \cellcolor[HTML]{DAE8FC}86.69 \\
 &
  \textbf{\textit{FT}} &
  \textbf{96.42} &
  05.47 &
  01.40 &
  00.40 &
  00.60 \\
 &
  \textbf{LwF} &
  96.38 &
  \textbf{47.98} &
  \textbf{11.67} &
  \textbf{00.36} &
  \textbf{00.76} \\
\multirow{-4}{*}{\textbf{DGCNN}~\cite{dgcnn}} &
  \textbf{Census} &
  {\ul \textbf{96.58}} &
  {\ul \textbf{72.37}} &
  {\ul \textbf{52.56}} &
  {\ul \textbf{36.27}} &
  {\ul \textbf{27.19}} \\
 &
  \cellcolor[HTML]{DAE8FC}\textbf{\textit{Joint}} &
  \cellcolor[HTML]{DAE8FC}96.84 &
  \cellcolor[HTML]{DAE8FC}95.41 &
  \cellcolor[HTML]{DAE8FC}90.94 &
  \cellcolor[HTML]{DAE8FC}88.01 &
  \cellcolor[HTML]{DAE8FC}87.23 \\
 &
  \textbf{\textit{FT}} &
  \textbf{96.67} &
  01.16 &
  00.35 &
  00.40 &
  00.32 \\
 &
  \textbf{LwF} &
  \textbf{96.67} &
  \textbf{52.14} &
  \textbf{04.17} &
  \textbf{00.40} &
  \textbf{00.89} \\
\multirow{-4}{*}{\textbf{PRA-Net}~\cite{PRA-NET}} &
  \textbf{Census} &
  {\ul \textbf{96.92}} &
  {\ul \textbf{72.31}} &
  {\ul \textbf{59.72}} &
  {\ul \textbf{47.61}} &
  {\ul \textbf{35.73}} \\ \hline \hline
\end{tabular}%
}
\end{table}

\begin{table*}[h]
\centering
\caption{Performance of different backbones on ModelNet40 considering one of the scenarios with 7 tasks modeled by PointCLIMB. We represent our findings by 1) best by \textbf{\uline{bold underline}} and 2) second best by \textbf{bold} accuracy values.}
\label{tab:task10}
\resizebox{\linewidth}{!}{%
\begin{tabular}{rccccccc}
\hline \hline
\textbf{Methods} &
  \textbf{10} &
  \textbf{5} &
  \textbf{5} &
  \textbf{5} &
  \textbf{5} &
  \textbf{5} &
  \textbf{5} \\ \hline
\textbf{Pointnet}~\cite{pointnet} &
  96.11 $\pm$ 1.29 &
  40.88 $\pm$ 9.36 &
  20.39 $\pm$ 5.63 &
  10.26 $\pm$ 2.29 &
  4.56 $\pm$ 1.19 &
  \textbf{5.59 $\pm$ 2.29} &
  4.47 $\pm$ 1.62 \\
\textbf{Pointnet++}~\cite{pointnetpp} &
  95.84 $\pm$ 2.17 &
  47.20 $\pm$ 8.86 &
  \textbf{22.61 $\pm$ 3.49} &
  \textbf{13.09 $\pm$ 5.74} &
  \textbf{6.79 $\pm$ 1.33} &
  4.82 $\pm$ 1.67 &
  \textbf{6.73 $\pm$ 4.80} \\
\textbf{DGCNN}~\cite{dgcnn} &
  {\ul \textbf{96.78 $\pm$ 0.73}} &
  \textbf{54.80 $\pm$ 13.58} &
  11.60 $\pm$ 4.42 &
  5.59 $\pm$ 0.97 &
  6.22 $\pm$ 0.64 &
  4.42 $\pm$ 1.17 &
  3.92 $\pm$ 2.25 \\
\textbf{PointMLP}~\cite{PointMLP} &
  95.04 $\pm$ 0.59 &
  44.40 $\pm$ 4.81 &
  13.36 $\pm$ 3.68 &
  5.72 $\pm$ 3.33 &
  4.64 $\pm$ 0.72 &
  5.39 $\pm$ 0.80 &
  3.60 $\pm$ 0.27 \\ \hline
\textbf{Ours (PRA-Net}~\cite{PRA-NET} \textbf{+ Census)} &
  \textbf{97.45 $\pm$ 1.86} &
  {\ul \textbf{69.49 $\pm$ 3.99}} &
  {\ul \textbf{49.92 $\pm$ 6.74}} &
  {\ul \textbf{36.19 $\pm$ 5.19}} &
  {\ul \textbf{33.47 $\pm$ 2.28}} &
  {\ul \textbf{24.04 $\pm$ 3.89}} &
  {\ul \textbf{21.96 $\pm$ 2.8}} \\ \hline \hline
\end{tabular}%
}
\end{table*}

~\begin{table*}[h]
\centering
\caption{Performance of different backbones on ModelNet40 considering one of the scenarios with uniform tasks modeled by PointCLIMB. We represent our findings by 1) best by \textbf{\uline{bold underline}} and 2) second best by \textbf{bold} accuracy values.}
\label{tab:task4}
\resizebox{\linewidth}{!}{%
\begin{tabular}{rcccccccccc}
\hline \hline
\textbf{Methods} &
  \textbf{4} &
  \textbf{4} &
  \textbf{4} &
  \textbf{4} &
  \textbf{4} &
  \textbf{4} &
  \textbf{4} &
  \textbf{4} &
  \textbf{4} &
  \textbf{4} \\ \hline 
\textbf{Pointnet}~\cite{pointnet} &
  96.87 $\pm$ 1.96 &
  32.81 $\pm$ 10.55 &
  17.18 $\pm$ 8.80 &
  9.25 $\pm$ 2.65 &
  3.45 $\pm$ 1.28 &
  2.18 $\pm$ 0.09 &
  \textbf{3.34 $\pm$ 1.18} &
  4.36 $\pm$ 2.29 &
  \textbf{4.01 $\pm$ 1.95} &
  3.52 $\pm$ 1.07 \\
\textbf{Pointnet++}~\cite{pointnetpp} &
  \textbf{97.18 $\pm$ 1.54} &
  \textbf{32.81 $\pm$ 13} &
  \textbf{21.45 $\pm$ 4.47} &
  8.33 $\pm$ 4.40 &
  5.80 $\pm$ 2.86 &
  5.33 $\pm$ 0.99 &
  3.09 $\pm$ 0.54 &
  \textbf{4.61 $\pm$ 0.46} &
  2.36 $\pm$ 1.14 &
  3.76 $\pm$ 1.71 \\
\textbf{DGCNN}~\cite{dgcnn} &
  {\ul \textbf{97.81 $\pm$ 2.15}} &
  50.12 $\pm$ 12.29 &
  20.88 $\pm$ 5.88 &
  7.66 $\pm$ 4.28 &
  5.06 $\pm$ 2.2 &
  2.46 $\pm$ 0.14 &
  2.84 $\pm$ 1.09 &
  4.46 $\pm$ 0.37 &
  3.24 $\pm$ 1.13 &
  2.13 $\pm$ 0.42 \\
\textbf{PointMLP}~\cite{PointMLP} &
  96.875 $\pm$ 0.45 &
  30.68 $\pm$ 7.69 &
  16.66 $\pm$ 6.80 &
  \textbf{9.5 $\pm$ 3.47} &
  \textbf{7.07 $\pm$ 1.19} &
  \textbf{5.65 $\pm$ 1.11} &
  2.30 $\pm$ 1.27 &
  4.2 $\pm$ 1.15 &
  2.11 $\pm$ 0.87 &
  \textbf{4.93 $\pm$ 2.33} \\ \hline
\textbf{Ours (PRA-Net}~\cite{PRA-NET} \textbf{+ Census)} &
  97.12 $\pm$  1.22 &
  {\ul \textbf{56.76 $\pm$ 11.47}} &
  {\ul \textbf{31.4 $\pm$ 5.88}} &
  {\ul \textbf{25.70 $\pm$ 5.26}} &
  {\ul \textbf{20.97 $\pm$ 3.88}} &
  {\ul \textbf{19.53 $\pm$ 5.53}} &
  {\ul \textbf{16.85 $\pm$ 2.27}} &
  {\ul \textbf{14.55 $\pm$ 3.73}} &
  {\ul \textbf{10.34 $\pm$ 3.98}} &
  {\ul \textbf{8.18 $\pm$ 2.41}} \\ \hline \hline
\end{tabular}%
}
\end{table*}

\noindent \textbf{Implementation Details} To train \textbf{PointCLIMB}, we used 1024 points and trained each task for 40 epochs using the Adam optimizer with a learning rate of 0.0001 and a batch size of 32. To dynamically vary the logits of the classification layer based on the number of arrived novel classes, we referred to the 3D-FSCIL\cite{chowdhury2022few} approach. We also utilized their class and label mapper for arrived novel data and modified the incremental data loader to make it exemplar-free. Our implementation of PointCLIMB was based on the PyTorch framework and ran on an NVIDIA Quadro GV100 32GB.\\

\noindent \textbf{Evaluation} of \textbf{PointCLIMB}, we used the ModelNet40 dataset and assessed the performance of different backbones. Specifically, we considered a scenario where the base task consists of 20 random classes, and each novel task contains 5 randomly chosen classes for Class-Incremental learning as depicted in Table~\ref{tab:all-loss}. To quantify the performance of different backbones, we compared naive fine-tuning, LwF \cite{li2017learning}, and Census. We evaluated PointCLIMB on different scenarios modeled by a veristic task sampler, which involved three scenarios. The first scenario had 20 random classes in the base task and 5 random classes in each novel task, with a total of 5 tasks. The results of this scenario are reported in Table \ref{tab:task20}. The second scenario involved 10 random classes in the base task and 5 random classes in each novel task, with a total of 7 tasks. The results of this scenario are also reported in Table \ref{tab:task20}. Lastly, we considered a scenario consisting of tasks with all uniform classes, where 4 classes were encountered in both the base and novel tasks, with a total of 10 tasks. The results of this scenario are reported in Table \ref{tab:task20}. The endless scenarios representing the actual paradigm of Class Incremental learning can be modeled using PointCLIMB. We benchmarked three possible scenarios of PointCLIMB for benchmarking, demonstrating the flexibility and effectiveness of our approach.\\

\noindent \textbf{Results and Discussion} As shown in Table \ref{tab:all-loss}, we provide upper (Joint) and lower bounds (Fine Tuning) on different backbone architectures. To evaluate the performance of LWF\cite{li2017learning} with different backbones, we report task-wise accuracies. Moreover, we introduce a novel variant of LWF loss, known as Census knowledge distillation loss, and observe a remarkable improvement in performance compared to the traditional LWF loss. Among all the backbone architectures, PRA-Net with Census Knowledge distillation loss outperforms others due to its capability to extract superior geometric features and use gated units to maintain stability. Thus, for modeling pragmatic scenarios, we recommend using PRA-Net and Census Knowledge Distillation \ref{eq:lcensus} as the superior baseline for 3D-EFCIL towards future research.

To evaluate the performance of the pragmatic scenarios modeled by the veristic task sampler described in Algorithm \ref{alg:veristic} on different backbones using LWF loss versus the suggested approach (PRA-Net + Census Knowledge Distillation), we conduct extensive ablation experiments with 5, 7, and 10 tasks in Tables~\ref{tab:task20}, \ref{tab:task10}, and \ref{tab:task4}, respectively. We run 3 experiments with different random seeds to assess the true response of backbones towards realistic settings sampled by PointCLIMB, and report the mean and standard deviation of all the backbones for better comprehensibility. Our results demonstrate that the suggested baseline outperforms all other backbones with all possible pragmatic settings. Apart from our recommended baseline, PointNet++~\cite{pointnetpp} performs the second-best in most scenarios and various incremental tasks. Moreover, we observe that all other architectures do not produce stable outcomes, unlike our proposed baseline that remains stable throughout the scenarios and incremental tasks. \\

\useunder{\uline}{\ul}{}
\begin{table*}[t]
\centering
\caption{Performance of different backbones on ModelNet40 considering one of the scenarios with 5 tasks modeled by PointCLIMB. We represent our findings by 1) best by \textbf{\uline{bold underline}} and 2) second best by \textbf{bold} accuracy values.}
\label{tab:task20}
\resizebox{\linewidth}{!}{%
\begin{tabular}{rccccc}
\hline \hline
\textbf{Methods}    & \textbf{20}          & \textbf{5}           & \textbf{5}           & \textbf{5}          & \textbf{5}          \\ \hline
\textbf{Pointnet}~\cite{pointnet}   & 95.25 $\pm$ 2.47          & 53.37 $\pm$ 5.56          & 8.92 $\pm$ 2.50           & 5.31 $\pm$ 1.70          & \textbf{6.31 $\pm$ 1.24} \\
\textbf{Pointnet++}~\cite{pointnetpp} & \textbf{96.83 $\pm$ 0.86} & \textbf{63.62 $\pm$ 8.56} & 6.49 $\pm$ 2.05           & \textbf{7.94 $\pm$ 1.58} & 4.39 $\pm$ 2.75          \\
\textbf{DGCNN}~\cite{dgcnn}      & 96.33 $\pm$ 1.47          & 58.75 $\pm$ 10.78         & \textbf{11.67 $\pm$ 5.73} & 7.48 $\pm$ 2.69          & 3.25 $\pm$ 0.96          \\
\textbf{PointMLP}~\cite{PointMLP}   & 95.50 $\pm$ 2.27          & 61.21 $\pm$ 8.62          & 6.74 $\pm$ 1.56           & 3.99 $\pm$ 0.27          & 2.93 $\pm$ 0.19          \\ \hline
\textbf{Ours (PRA-Net}~\cite{PRA-NET} \textbf{+ Census)} &
  {\ul \textbf{96.92 $\pm$ 0.74}} &
  {\ul \textbf{67.30 $\pm$ 5.32}} &
  {\ul \textbf{55.17 $\pm$ 7.51}} &
  {\ul \textbf{45.12 $\pm$ 5.61}} &
  {\ul \textbf{30.18 $\pm$ 7.31}} \\ \hline \hline
\end{tabular}%
}
\end{table*}

\subsection{Limitations}
    Despite its potential, 3D exemplar-free class incremental learning has some limitations that must be considered. One of the main challenges is the need for large amounts of high-quality data to train and evaluate the model. This can be particularly difficult in 3D environments, where obtaining and processing data can be time-consuming and expensive. Current backbone that we test are not robust to noise so this is major limitation. Additionally, 3D exemplar-free class incremental learning may face limitations when dealing with complex and highly variable objects, where there may be significant intra-class, inter-class variation with-in and among incremental tasks. This can lead to difficulty in accurately classifying these objects and may require more specialized models or additional training data.
    
    Although there are some limitations to 3D exemplar-free class incremental learning, we are confident that our study provides valuable insights within point cloud environments. We are optimistic that our research will inspire further investigation into these limitations, ultimately leading to the development of more resilient and efficient methods for 3D exemplar-free class incremental learning.

\section{Conclusions}
In this paper, we pioneered into the uncharted territory of exemplar-free class incremental learning on point clouds (3D-EFCIL) and presented a pragmatic experimental setup that mirrors real-world continual learning scenarios. Our investigation into various Point cloud classification backbones yielded encouraging results on the ModelNet40 dataset for 3D exemplar-free class incremental learning. Our analysis revealed that backbones with intra-inter local neighbourhood relation awareness significantly outperformed global topology-based and local neighbourhood-based methods on 3D-EFCIL. Furthermore, we explored how the use of weighted knowledge distillation loss (census) could alleviate the issue of catastrophic forgetting in 3D-EFCIL. We anticipate that our work will inspire further research in this field, leading to the emergence of more effective and robust methods for few-shot learning in point clouds.

\section{Broader Impact}
    The main goal of this research is to establish realistic benchmarks for 3D exemplar-free class incremental learning, and explore the optimal feature extractors / backbone, and network design choices for knowledge distillation loss in this context. 
    
    3D exemplar-free incremental learning has the potential to generate significant and far-reaching impacts in various fields. By enabling machines to learn new object categories without prior examples, exemplar-free incremental learning can facilitate real-time learning and adaptation in dynamic scenarios, making it ideal for robotics, computer vision, and artificial intelligence. This research can improve the efficiency of machine learning algorithms by reducing computational costs and enable autonomous vehicles, robots to operate more safely and efficiently. Additionally, exemplar-free incremental learning can enhance quality control in manufacturing processes by detecting defects and ensuring product specifications are met. Overall, this research has the potential to transform several industries, paving the way for innovative solutions and improving efficiency, accuracy, and safety.

{\small
\bibliographystyle{ieee_fullname}
\bibliography{main}
}

\end{document}